\documentclass[runningheads]{llncs}

\usepackage[utf8]{inputenc}
\usepackage[T1]{fontenc}
\usepackage{graphicx}
\usepackage{float}
\usepackage{amsmath}
\usepackage{amssymb}
\usepackage{pgfplots}
\pgfplotsset{compat=1.18}
\usepackage{tikz}
\usetikzlibrary{arrows.meta, positioning, shadows.blur}
\newcommand{\arm}[1]{\includegraphics{arm-snippets/#1}}
\usepackage{hyperref}

\newcommand{\twodigits}[1]{\ifnum#1<10 0\fi\number#1}

\graphicspath{{section_curvature/}{section_layout/}{section_pipeline/}{section_intro/}{section_datasetannotation/}{section_db_map/}{data/}}

\begin{document}

\title{Mapping Armenian Paris: Extracting and Geocoding Commercial Advertisements from the 20th-Century Diaspora Press}
\titlerunning{Mapping Armenian Paris}

\author{Chahan Vidal-Gorène\inst{1,2} \and Seda Kirakosyan\inst{1,3} \and Edita Matevosyan\inst{1,3}}

% \author{Chahan Vidal-Gorène\inst{1}\orcidID{} \and
% Second Author\inst{2}\orcidID{} \and
% Third Author\inst{3}\orcidID{}}

\authorrunning{C. Vidal-Gorène et al.}

\institute{%
Calfa, Paris, France \and
École nationale des chartes, PSL University, Paris, France \and
Université française en Arménie (UFAR), Yerevan, Armenia \\
\email{surname.name@calfa.fr}
}

\maketitle

\begin{abstract}
This paper presents an end-to-end, IIIF-based pipeline that turns the digitised
Armenian press of France into an interactive map of the 20th-century
Parisian Armenian commercial community. On each page, commercial
advertisements are located, read, and parsed into structured records,
which are then geocoded and placed on the map. Western Armenian is
under-resourced and unsupported by off-the-shelf layout and OCR
models, so the pipeline uses vision--language models (VLMs) as a
data-bootstrapping strategy: they produce usable structured records
at a scale hand annotation could not reach, and stay reliable on the
strongly curved scans where conventional line-level CRNN OCR breaks down. The
contribution includes a $500$-page Western Armenian press corpus with
$3{,}270$ advertisement-level annotations, a Label Studio template
that captures detection and semantic fields in a single annotation
pass, and a reproducible workflow transposable to other
under-resourced historical corpora. More broadly, the work shows that VLM-driven data
bootstrapping is an effective lever for under-resourced historical
languages such as (Western) Armenian.

\end{abstract}

\section{Introduction and context of this research}
\label{sec:intro}

The Armenian press spans more than two centuries -- from
\emph{Azdarar} (Madras, 1794), through a 19th-century golden age in
Constantinople, Tiflis and Moscow, to a post-1915 redirection toward
the Middle Eastern, European and American diasporas. In France, the
diaspora produced a dense ecosystem of Western Armenian dailies and
weeklies, particularly between 1920 and 1940, whose
\emph{commercial advertisements} form a fine-grained record of the
Parisian Armenian commercial community of the period. Most of these periodicals were short-lived and
folded before 1940; the daily \emph{Haratch}, published in Paris
until 2009, is the main long-running exception and remains the
central resource for the era. The \textbf{MAP} project
(\emph{Mapping Armenian Paris}) sets out to turn this nearly
untapped record into an interactive cartography of the Armenian
commercial community of 20th-century Paris, drawing on the
Armenian press of France digitised by ARAM--BULAC, the BnF, the
National Library of Armenia (NLA), and by Calfa from the
Samuelian collection\footnote{Digital archives of the Armenian press (primarily scans rather than plain text) are accessible through four main institutions. The National Library of Armenia (NLA) [\href{https://tert.nla.am/}{tert.nla.am}] constitutes the largest database compiling and digitizing these records. Other key repositories include the Pan-Armenian Digital Library ARAR [\href{https://arar.sci.am/}{arar.sci.am}], supporting in particular digitization efforts in Vienna (Mekhitarists) and BULAC [\href{https://www.bulac.fr/}{bulac.fr}] (with ARAM association). These combined efforts reflect a broader initiative supported by the Calouste Gulbenkian Foundation to digitize and preserve the Western Armenian press.}.

% \begin{figure}[!htbp]
% \centering
% \includegraphics[width=\linewidth,keepaspectratio]{scanner}
% \caption{Large-format imaging rig built for the on-site
%   digitisation of the Samuelian press collection: a
%   Raspberry~Pi~4 (top right) drives a 64\,Mpx autofocus Arducam
%   camera, with three LED panels. Image capturing and framing are AI-assisted in Python via Ultralytics and the pycamera package.}
% \label{fig:scanner}
% \end{figure}

\paragraph{Additional context:} Many titles relevant to MAP never reached Gallica or the
NLA: they escaped French legal deposit and survived only as physical
copies at the \emph{Samuelian} oriental bookstore (Paris,
1930--2016), closed to the public for nearly a decade, where the
collection had been stored under conditions unfit for long-term
preservation. Between 2024 and
2025, with the support of the Calouste Gulbenkian Foundation and
in coordination with the FSL of NASRA, the BnF, and the Samuelian familly, Calfa
inventoried and digitised on site this collection of 20th-century Parisian Armenian
press ($15{,}000$ pages)\footnote{Due to bookstore constraints and the fragile, oversized nature of the pages, standard flatbed scanners were unsuitable. Calfa developed a custom large-format imaging rig utilizing a Raspberry~Pi~4, a 64,Mpx Arducam sensor on a central boom, and three LED panels on 360° pivots. AI-assisted focus and framing were implemented in Python via Ultralytics for document and page detection following~\cite{grigoryan2025automated}. The resulting corpus will be integrated into Gallica (BnF Digital Library) and ingested into the MAP pipeline described herein.}.

\section{Related works}
\label{sec:related}

\paragraph{Historical document OCR and layout.} Historical
newspaper digitisation has long combined Document Layout Analysis
with OCR to cope with irregular layouts, typographical variation
and physical degradation~\cite{fleischhacker2025enhancing}. Recent
end-to-end Vision--Language Models bypass this detect-then-recognise
cascade by jointly localising and transcribing text on the raw page
image: the Qwen family and its fine-tuned derivatives
(CHURRO~\cite{semnani2025churro}, Chandra OCR),
GLM-OCR~\cite{duan2026glm} and
DeepSeek~OCR~\cite{wei2025deepseek,wei2026deepseek} are recent
examples of this shift toward single-model, end-to-end reading.

\paragraph{Geohistorical GIS.} On the spatial side, projects such as
SoDUCo~\cite{abadie2022benchmark,abadie2023soduco} and Paris Time
Machine~\cite{mermet2026galligeo,mermet2024developing,pinol2024adresses}
exploit OCR-processed trade directories and printed maps to
reconstruct urban evolution~\cite{gravier2024typology}, turning raw
text into geohistorical knowledge at city scale. These pipelines,
however, start from sources already in a modern, well-resourced
language and already OCR-processed to a usable standard.

\paragraph{Under-resourced languages and LLM-as-annotator
strategies.} These OCR and VLM advances remain biased toward
high-resource languages and modern Latin scripts, so under-resourced
languages cannot leverage modern LLMs and VLMs
directly~\cite{zhong2024opportunities,vidal2026under}. For Armenian
specifically, recent work has largely answered the text-recognition
task itself, both handwritten and
printed~\cite{VidalGorene2023,vidal2026-newspaper,vidal2025armenian},
shifting the bottleneck from recognition to the creation of
structured training and evaluation data. Using LLMs as annotators for low-resource historical
languages~\cite{vidal2026under} marks a major turning point: it
sharply lowers the cost of producing annotated corpora and offers a
credible path out of the under-resourced regime. The present paper
extends that line of work one level up, from isolated text
recognition to the page and advertisement level, and couples it for
the first time with a full geocoding pipeline on the Armenian press
of France.

\section{Dataset and annotations}
\label{sec:dataset}

The dataset 
contains $500$ pages of Armenian press of France -- $66$ of them
reused from~\cite{vidal2026-newspaper} and $434$ newly compiled
for the present work. Of these, $430$ pages come from the Parisian
daily \emph{Haratch}, harvested from the BULAC \emph{Bina} digital
library; the remaining $70$ pages cover seven early-20th-century
Armenian-language titles harvested from \emph{Gallica} (BnF).
Table~\ref{tab:dataset} summarises this distribution. For \emph{Haratch}, 294 of the 430 pages contain full-page advertisements; the remaining 136 pages are sampled to capture ads embedded in the editorial layout or do not contain any ads. In total the corpus carries $3{,}270$
region-level annotations split between $994$ framed ads
(\texttt{ad\_with\_frame}) and $2{,}276$ borderless ads
(\texttt{ad\_without\_frame}).

\begin{table}[b]
\centering
\caption{Composition of the $500$-page advertisement corpus
  ($3{,}270$ region annotations: $994$ framed, $2{,}276$
  borderless).}
\label{tab:dataset}
\setlength{\tabcolsep}{6pt}
\renewcommand{\arraystretch}{1.05}
\footnotesize
\begin{tabular}{l l l r}
\hline
Title & Digital library & Years & Pages \\
\hline
\emph{Haratch}        & Bina (BULAC)  & 1925--1997   & 430 \\
\emph{Aiguillon}      & Gallica (BnF) & 1919         &   6 \\
\emph{Armenia}        & Gallica (BnF) & 1903--1923   &  11 \\
\emph{Gaghapar}       & Gallica (BnF) & 1916--1917   &   6 \\
\emph{L'Arménie}      & Gallica (BnF) & 1903--1905   &   8 \\
\emph{Le Foyer}       & Gallica (BnF) & 1928--1932   &   9 \\
\emph{Mer-Oughine}    & Gallica (BnF) & 1931--1932   &  15 \\
\emph{Nor Guiank}     & Gallica (BnF) & n.d.         &  15 \\
\hline
\textbf{Total}        &               &              & \textbf{500} \\
\hline
\end{tabular}
\end{table}

\begin{figure}[t]
\centering
\includegraphics[width=\linewidth,keepaspectratio]{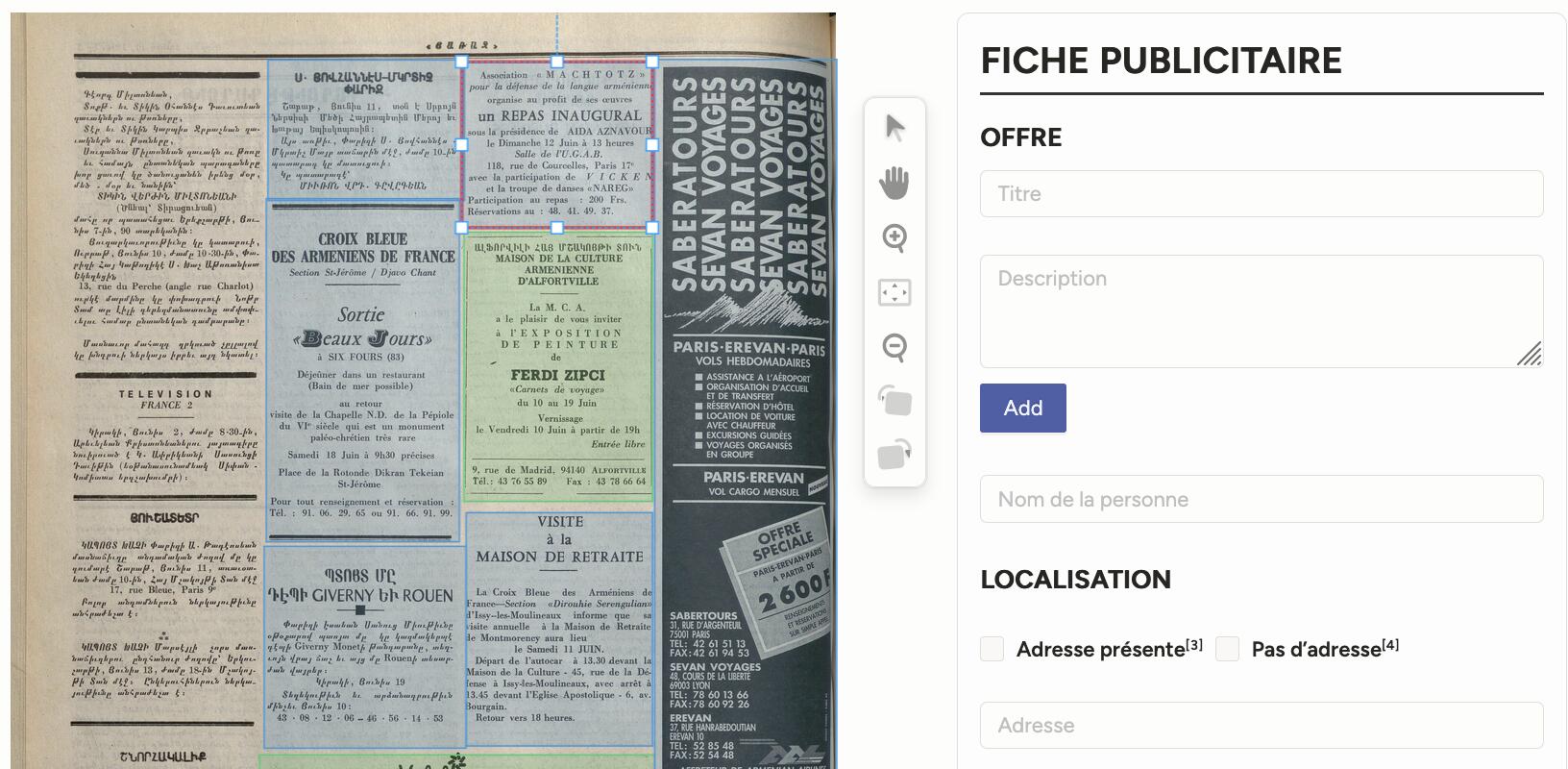}
\caption{Custom Label Studio interface: rectangle labels for
  advertisement detection on the left; the hierarchical semantic
  form on the right is only revealed when a region is selected.}
\label{fig:labelstudio}
\end{figure}

\paragraph{Annotation interface.} All $3{,}270$ regions were
produced with a custom \textbf{Label Studio} template that
combines, in a single schema and a single annotation pass, a
\emph{detection layer} (rectangle labels \texttt{with-frame} /
\texttt{without-frame}) and a per-region \emph{semantic layer}
(structured offer fields, location, contact, languages, hierarchical
topic categories, visuals). A single export format then supports training an end-to-end
multi-modal VLM, training the staged
detector\,+\,OCR\,+\,parser pipelines benchmarked here, or
prototyping new experiments without re-annotation. The
template is released with the dataset as a reusable Label Studio
configuration (see Data Availability). Figure~\ref{fig:labelstudio}
reproduces the interface; Table~\ref{tab:schema} in
Appendix~\ref{app:prompts} lists the JSON fields exported per
region.

\begin{figure}[t]
\centering
\includegraphics[width=\linewidth,height=0.32\textheight,keepaspectratio]{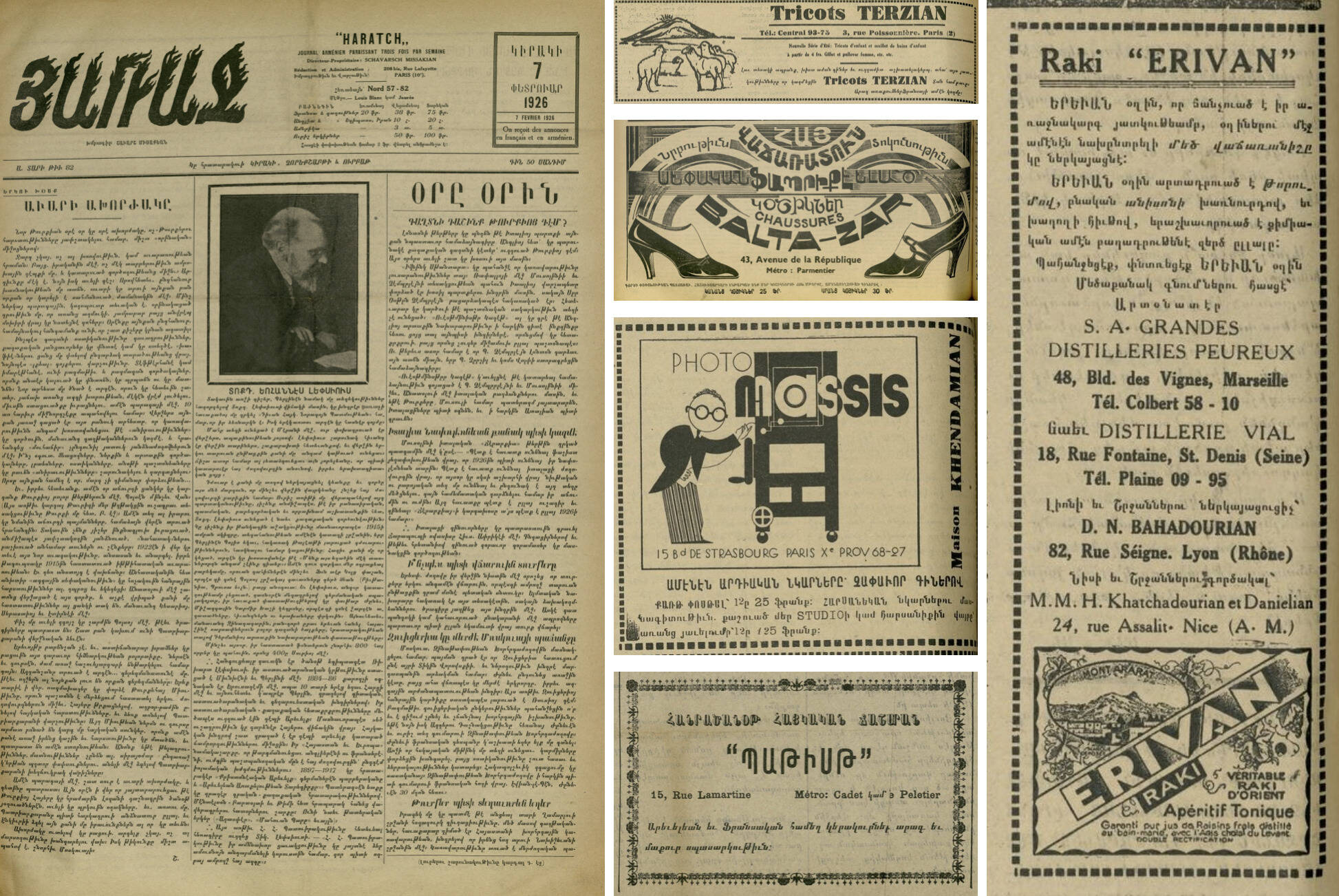}
\caption{Example of layout and advertisements. Details in Table~\ref{tab:dataset}. Haratch newspaper, 1926, digitized by BULAC and ARAM, available on Bina.}
\label{fig:corpus}
\end{figure}

\section{End-to-end pipeline}
\label{sec:pipeline}

\begin{figure}[!htbp]
\centering
\includegraphics[width=\linewidth,height=0.85\textheight,keepaspectratio]{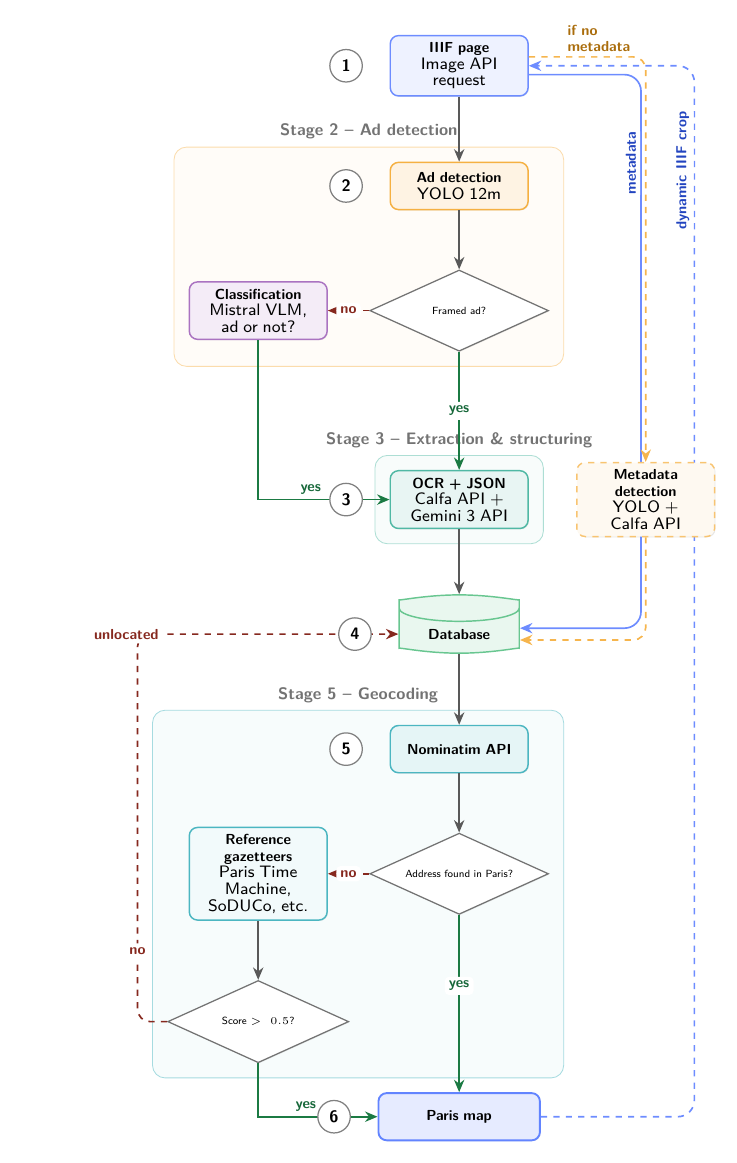}
\caption{End-to-end pipeline. Solid arrows mark the default
  flow, dashed arrows the fallback branches: metadata recovery
  when the IIIF manifest is incomplete, the gazetteer lookup when
  Nominatim cannot resolve the address, and the dynamic-crop
  feedback from the Paris map. Numbered badges follow the
  walk-through in the text.}
\label{fig:pipeline}
\end{figure}

Figure~\ref{fig:pipeline} sketches the
six stages. (1)~The input is a IIIF-served newspaper page, fetched
on demand through the IIIF Image~API. (2)~A YOLO12m detector (object detection)
locates candidate advertisements on the page. Framed detections
(\texttt{ad\_with\_frame}) are accepted directly; borderless
detections are routed through a Mistral-family vision--language
classifier that decides whether the crop is an advertisement at all
-- negative verdicts are dropped, positive verdicts rejoin the main
flow. (3)~The retained
crops are sent through an extraction stage that combines the
Calfa OCR API for line transcription on regular crops with the
Gemini~3~Flash API, which serves two roles: OCR on strongly
curved crops where line-level CRNNs collapse
(Section~\ref{sec:ocr}), and ad-level structured field extraction
(title, description, address components, telephone, languages,
topics; see Figure~\ref{fig:gemini_prompt} in
Appendix~\ref{app:prompts}).
(4)~The structured records are persisted to the database. In
parallel, page-level metadata are either read from the IIIF
manifest (default branch, solid arrow on the right of
Figure~\ref{fig:pipeline}) or, when the manifest is incomplete,
recovered by a second YOLO detector that locates the bibliographic
strip and feeds it to the Calfa OCR API (dashed branch).
(5)~Each extracted address is geocoded through the Nominatim API.
Addresses returned directly in Paris are accepted; the others fall
back to reference gazetteers (Paris Time Machine, SoDUCo, and
similar historical sources) and are retained only above a
similarity score of $0.5$. Addresses that still cannot be resolved
are stored in the database as a separate ``unlocated'' list,
available for later manual curation. (6)~Successfully geocoded
advertisements feed an interactive Paris map view, which closes
the loop by issuing on-demand IIIF crops back to the original
pages, so that any retrieved ad can be inspected in its source
context without a local image cache.

\paragraph{IIIF as the backbone.} Every page and crop is requested
as a IIIF Image~API URL, so no high-resolution scan is downloaded or
cached locally. IIIF is used only by the live deployed pipeline for
on-demand fetching at scale: the released $500$-page benchmark
corpus (Section~\ref{sec:dataset}) is distributed as plain images
with JSON annotations and needs no IIIF tooling to reuse.

\paragraph{Current limitations.} The pipeline assumes IIIF-served
sources. Smaller archives that expose incomplete manifests are
handled by the stop-gap YOLO-based metadata-recovery branch
(dashed path in Figure~\ref{fig:pipeline}); fully non-IIIF
endpoints are out of scope and left to future work
(Section~\ref{sec:conclusion}). A second limitation is
reproducibility: the pipeline currently relies on external paid
APIs (Mistral, Gemini, Calfa~OCR), so reusing it on
another corpus requires budgeting for per-call API costs; the
production transition to a locally-hosted, fine-tuned VLM
(Section~\ref{sec:conclusion}) would remove this dependency.

\section{Advertisement layout detection}
\label{sec:layout}

\paragraph{Task and models.} Advertisement localisation on the
\emph{Haratch} pages is framed as a two-class detection task,
with classes \texttt{ad\_with\_frame} (visible ruled border) and
\texttt{ad\_without\_frame} (borderless commercial blocks set in
running text). LayoutLM-style multimodal detectors~\cite{xu2020layoutlm} are excluded:
they require word-level bounding boxes paired with OCR tokens, no
pretrained checkpoint covers Armenian (an under-resourced
language), and the corpus is annotated at region level only.
Vision-only detectors are therefore benchmarked:
\textbf{YOLOv8(m)}, \textbf{YOLOv12m}, \textbf{YOLO26m} and the
transformer-based \textbf{RT-DETR-L}, all initialised from
COCO-pretrained weights.\footnote{All runs use input size $1024$ and
$100$ epochs. Early-stopping patience is $20$, with AMP. The optimiser
is auto-selected AdamW ($\mathrm{lr}_0\!=\!10^{-2}$, momentum $0.937$,
weight decay $5\!\cdot\!10^{-4}$, $3$ warm-up epochs). The augmentation
stack is shared: mosaic, fliplr $0.5$, RandAugment, HSV jitter, random
scaling $\pm 0.5$, random erasing $0.4$. Only the base weights, the
batch size (6--12, GPU-memory constrained) and the training set vary.
Three dataset sizes are used: \textsf{ad\_detection100} (20 val
images), \textsf{ad\_detection400} (60 val), \textsf{ad\_detection500}
(75 val).}

\begin{table}[t]
\centering
\caption{Layout detector benchmark on the advertisement detection
  test split. \textbf{Bold} marks the best value per metric; mAP50
  and mAP50-95 do not agree on a winner among the
  \textsf{ad\_detection400} runs (Exp~3--6, see Results).}
\label{tab:layout_main}
\setlength{\tabcolsep}{3pt}
\renewcommand{\arraystretch}{1.05}
\resizebox{\linewidth}{!}{%
\footnotesize
\begin{tabular}{c l l r r r r r r}
\hline
Exp & Model & Dataset & Val & Params (M) & P & R & mAP50 & mAP50-95 \\
\hline
2 & RT-DETR-L  & ad\_detection100  & 20 & 32.0 & 0.833 & 0.835 & 0.912 & 0.816 \\
3 & YOLOv8m    & ad\_detection400  & 60 & 25.8 & 0.866 & 0.894 & 0.917 & \textbf{0.842} \\
4 & RT-DETR-L  & ad\_detection400  & 60 & 32.0 & 0.863 & 0.918 & 0.911 & 0.841 \\
5 & YOLOv12m   & ad\_detection400  & 60 & 20.1 & 0.891 & 0.880 & \textbf{0.927} & 0.840 \\
6 & YOLO26m    & ad\_detection400  & 60 & 20.4 & 0.851 & 0.901 & 0.911 & 0.836 \\
7 & YOLOv12m   & ad\_detection500  & 75 & 20.1 & 0.872 & 0.911 & 0.919 & 0.819 \\
\hline
\end{tabular}}
\end{table}

\paragraph{Results.} From \textsf{ad\_detection400} onwards, the four detectors are statistically tied: mAP50 spans $[0.911, 0.927]$ and mAP50:95 spans $[0.836, 0.842]$, a $0.6$-point spread, and the two metrics disagree on a winner. \textbf{YOLOv8m} (Exp~3) leads on mAP50:95 ($0.842$), ahead of \textbf{RT-DETR-L} (Exp~4, $0.841$) and \textbf{YOLOv12m} (Exp~5, $0.840$), while \textbf{YOLOv12m} leads on mAP50 ($0.927$ vs.\ $0.917$/$0.911$). Precision/recall split the same way: YOLOv12m has the best precision ($0.891$) but lowest recall ($0.880$); RT-DETR-L has the best recall ($0.918$) at lower precision ($0.863$). Since accuracy cannot pick a winner, \textbf{YOLOv12m (Exp~5)} is chosen instead for cost: fewest parameters ($20.1$\,M vs.\ $20.4$/$25.8$/$32.0$\,M for YOLO26m/YOLOv8m/RT-DETR-L) and fewest GFLOPs ($67.1$ vs.\ $67.9$/$78.7$/$103.4$).
Per-class results reveal the real difficulty, regardless of detector: \texttt{ad\_with\_frame} is essentially saturated (mAP50 $\in[0.975, 0.994]$, mAP50-95 $\in[0.932, 0.956]$ across Exp~3--7), while \texttt{ad\_without\_frame} trails by $15$--$25$ points on both metrics (Exp~5: mAP50 $0.861$ / mAP50-95 $0.724$ vs.\ $0.994$ / $0.956$ for framed ads), since borderless blocks resemble ordinary editorial text without OCR context. Expanding training data from $400$ to $500$ pages does not close this gap ($0.927\!\rightarrow\!0.919$ mAP50): the added pages mostly contribute more borderless instances — a class a pure-vision detector cannot disambiguate.

\paragraph{Ad-verification cascade.} To suppress residual \texttt{ad\_without\_frame} false positives, each borderless YOLO crop is sent to a Mistral-family vision--language model (OpenRouter) using the prompt in Figure~\ref{fig:vlm_prompt} (Appendix~\ref{app:prompts}): an early-\textsc{No} short-circuit on internal credits, broad \textsc{Yes} cues for ad signals, and a default-\textsc{Yes} tie-break for ambiguous cases. \textsc{Yes} verdicts are kept, \textsc{No} dropped, API failures flagged for review. To test model-generality, six configurations are compared: \textbf{Ministral~3B}, \textbf{Mistral~Medium}, and \textbf{Mistral~Large}, each zero-shot and with $7$ few-shot examples added to the prompt ($3$ framed, $3$ frameless, $1$ non-ad, each labelled with its expected verdict; both variants in Figure~\ref{fig:vlm_prompt}, Appendix~\ref{app:prompts}).

\begin{table}[t]
\centering
\caption{Ad-verification cascade: three
  Mistral-family models, zero-shot and with $7$ few-shot in-context
  examples (Appendix~\ref{app:prompts}). Pipeline metrics combine
  the verifier's verdict with the upstream YOLOv12m detector and the
  framed detections, which bypass the cascade. \textbf{Bold} marks
  the best value per column; $\dagger$ marks the configuration
  retained in production.}
\label{tab:layout_cascade}
\setlength{\tabcolsep}{5pt}
\renewcommand{\arraystretch}{1.1}
\footnotesize
\begin{tabular}{l l r r r r r r}
\hline
Model & Few-shot & Yes & No & Lost ads & Rec. & Prec. & F1 \\
\hline
YOLOv12m alone   & --  & --    & --  & --  & 0.884          & 0.897          & 0.890 \\
\hline
Ministral-3B     & no  & 2,299 & 457 & 249 & 0.761          & 0.926          & 0.835 \\
Ministral-3B     & yes & 2,048 & 709 & 556 & 0.685          & \textbf{0.936} & 0.791 \\
Mistral Medium   & no  & 2,421 & 336 & 222 & 0.804          & 0.929          & 0.862 \\
Mistral Medium$^\dagger$ & yes & 2,467 & 290 & 191 & \textbf{0.818} & 0.927          & \textbf{0.869} \\
Mistral Large    & no  & 2,374 & 311 & 191 & 0.787          & 0.927          & 0.851 \\
Mistral Large    & yes & 2,482 & 274 & 191 & 0.815          & 0.919          & 0.864 \\
\hline
\end{tabular}
\end{table}

\paragraph{Cascade impact.} Table~\ref{tab:layout_cascade} reports
all six configurations. Few-shot examples are not a universal gain:
they raise F1 for Mistral~Medium ($0.862\!\rightarrow\!0.869$) and
Mistral~Large ($0.851\!\rightarrow\!0.864$), but lower it for
Ministral~3B ($0.835\!\rightarrow\!0.791$, the worst of the six),
which over-rejects once examples are added ($709$ \textsc{No}
verdicts vs.\ $457$ zero-shot). The single best metric is also not a
safe criterion: Ministral-3B few-shot has the highest precision
($0.936$) only because it rejects the most genuine ads along with
the false positives. \textbf{Mistral~Medium, few-shot} is retained:
highest recall ($0.818$) and F1 ($0.869$) of the six, at the same
precision as the rest of the table ($0.927$).

Mistral~Medium few-shot shows a clear gain, not a
trade-off between two weaknesses: F1 rises from $0.835$ to $0.869$,
the recall the cascade costs against YOLOv12m alone is nearly halved
($12.3$~pts, $0.884\!\rightarrow\!0.761$, down to $6.6$~pts,
$0.884\!\rightarrow\!0.818$), and the precision gain is unchanged
($+2.9\!\rightarrow\!+3.0$~pts, $0.897\!\rightarrow\!0.927$). On this
$400$-page split, the $6.6$-point recall cost means $191$ genuine
advertisements are silently absent from the final corpus, with no
indication that anything was missed; the $3.0$-point precision gain
means $99$ fewer non-ads -- ordinary editorial text mistaken for a
borderless ad -- are wrongly kept in it, which would otherwise
inflate every downstream count (commerce density, language mix,
business type) with content that is not an advertisement at all.

The remaining $6.6$-point recall cost has an identifiable cause: of
the $290$ crops Mistral~Medium rejects, only $99$ ($34\,\%$) are
genuine non-ads, so two out of three \textsc{No} verdicts are
themselves wrong. These errors are concentrated in one class: $221$
of $1{,}607$ frameless ads are dropped ($13.8\,\%$) against $1$ of
$865$ framed ads ($0.1\,\%$). This is the same class already identified as the
harder one for the detector (Results above), not a separate failure
introduced by the verifier.

\section{Text recognition results}
\label{sec:ocr}

\paragraph{Models.} Seven OCR systems are benchmarked on the
advertisement test set described in Section~\ref{sec:dataset}: (i)
\textbf{Gemini 3~Flash}, a vision--language model accessed through OpenRouter~\cite{google2026gemini3flash}; (ii) \textbf{Qwen2.5-VL~72B-Instruct} and (iii)
\textbf{Qwen3-VL~32B-Instruct}, two open-weight vision--language
models; (iv) the
\textbf{default Armenian model of Tesseract~5} (CRNN+CTC); (v) a
\textbf{Tesseract~5 model trained by Calfa} on Armenian
newspapers~\cite{vidal2026-newspaper} (CRNN+CTC)\footnote{\url{https://github.com/calfa-co/hye-tesseract}}; (vi) the
production \textbf{Calfa~OCR}, a generic CRNN-based recognizer for
Armenian (handwritten and printed)~\cite{vidal2025armenian}; and (vii) \textbf{CRAFT~+~Tesseract~(Calfa)},
which replaces Tesseract's own page segmenter with
CRAFT~\cite{baek2019character} for text detection ahead of the same
Calfa-trained recognizer as (v). Reported below are both their
performance on the unaltered test set and their behaviour under
increasing synthetic page curvature, the most frequent geometric
defect of the scans.
\textbf{PaddleOCR} is not included: its released models do not cover
Armenian script, and training one is not realistic given the size of
our annotated corpus.

\paragraph{Curvature simulation.} The spine deformation typical of
bound newspapers is reproduced with a closed-form remapping. Let
$I:[0,W]\!\times\![0,H]\!\to\!\mathbb{R}^{3}$ be the source page and
$x_0 = \lfloor 0.4\,W \rfloor$ the abscissa at which the curvature
begins. For every column $x \ge x_0$, define the parabolic progress
ratio $t(x) = \bigl(\tfrac{x - x_{0}}{W - x_{0}}\bigr)^{2} \in [0,1]$
and, with three coupled parameters $(\alpha,\beta,\gamma)$:
\begin{align*}
  x'(x,y) &= x \;+\; t(x)\,(W - x_{0})\,\alpha , \\
  y'(x,y) &= y \;-\; t(x)\,\beta
                 \;+\; \tfrac{y - H/2}{H/2}\,t(x)\,\gamma .
\end{align*}
The first equation produces a horizontal compression of the page
towards the spine, while the second combines a vertical dip with a
perspective pinch around mid-height. The deformed image is obtained
by bilinear back-warping
$I'(x,y) = I\!\bigl(x'(x,y), y'(x,y)\bigr)$ with a white background.
For each advertisement in the test set, ten curvature levels
$i\in\{1,\dots,10\}$ are generated by linearly interpolating
$(\alpha,\beta,\gamma)$ from $(0,0,0)$ at $i=1$ to
$(0.65,\,250,\,125)$ at $i=10$, so that $i=1$ leaves the source flat
and $i=10$ matches the strongest curvature observed on the
production scans.

\paragraph{Results.} Figure~\ref{fig:curvature} plots the CER trend
with the segmentation-collapse threshold and one thumbnail per
intensity; Table~\ref{tab:curvature_scores} gives the exact CER
\emph{and} WER for every system and intensity. On the
unaltered test set ($i\!=\!1$), Calfa~OCR has the best score
($4.0\,\%$ CER~/~$23.4\,\%$ WER), closely followed by
CRAFT~+~Tesseract~(Calfa) ($5.0\,\%$~/~$24.4\,\%$); Gemini~3~Flash
is competitive ($6.3\,\%$~/~$40.4\,\%$, see the WER discussion below), the two Qwen
models trail at $33$--$37\,\%$ CER, and the two pipelines relying on
Tesseract's own page segmenter are already weaker ($\approx\!21\,\%$
CER) because of layout errors on the multi-block advertisement
crops.

Under increasing curvature ($i\!\geq\!2$), two regimes emerge.
Tesseract's own segmenter produces axis-aligned boxes that cannot
follow curved baselines past intensity~$\approx\!5$: CER crosses
$50\,\%$ between intensities~5 and~6 and exceeds $70\,\%$ from
intensity~7, for both the default and the Calfa-trained recognizer
-- once detection fails, recognizer quality stops mattering.
Every system that instead follows the curved text rather than
cropping it with an axis-aligned box stays well below $20\,\%$ CER
throughout: Calfa~OCR, whose line detector extracts each line as a
polygonal region following the curved baseline ($\leq\!14.2\,\%$);
CRAFT~+~Tesseract~(Calfa), which swaps in CRAFT for detection only
($\leq\!16.9\,\%$); and Gemini~3~Flash, which bypasses line
detection by reading the whole page at once ($\leq\!7.6\,\%$). The
two Qwen models are likewise insensitive to the deformation itself
but plateau at a higher floor ($22$--$49\,\%$, see the qualitative
analysis below). CRAFT~+~Tesseract~(Calfa) is the clearest evidence
that curvature is a \emph{detection}, not recognition, bottleneck:
it pairs CRAFT with the \emph{same} recognizer as the collapsing
Tesseract~(Calfa) pipeline, and swapping only the detector lowers
CER at $i\!=\!10$ from $79.5\,\%$ to $16.9\,\%$ and even gives it the
best WER of all seven systems at intensities~3--5. As a fully open,
non-generative pipeline nearly matching the proprietary Calfa~OCR
engine under curvature, it is a promising lightweight alternative to
a full VLM call. Gemini~3~Flash is retained for the strongly curved
subset and Calfa~OCR for the bulk of the corpus.

\begin{table}[t]
\centering
\caption{Average CER and WER (\%) on the advertisement test set.
  Row $i\!=\!1$ corresponds to the unaltered test set (general OCR
  results); rows $i\!=\!2,\dots,10$ apply the synthetic curvature of
  Section~\ref{sec:ocr} with $(\alpha,\beta,\gamma)$ linearly
  interpolated between $(0,0,0)$ at $i\!=\!1$ and $(0.65,250,125)$ at
  $i\!=\!10$. \textbf{Bold} marks the best system(s) per row and metric.}
\label{tab:curvature_scores}
\setlength{\tabcolsep}{3pt}
\renewcommand{\arraystretch}{1.05}
\scriptsize
\resizebox{\textwidth}{!}{%
\begin{tabular}{c|cc|cc|cc|cc|cc|cc|cc}
\hline
 & \multicolumn{2}{c|}{Tesseract\,5 (default)}
 & \multicolumn{2}{c|}{Tesseract\,5 (Calfa)}
 & \multicolumn{2}{c|}{CRAFT+Tess.\,(Calfa)}
 & \multicolumn{2}{c|}{Calfa\,OCR}
 & \multicolumn{2}{c|}{Gemini\,3 Flash}
 & \multicolumn{2}{c|}{Qwen2.5-VL\,72B}
 & \multicolumn{2}{c}{Qwen3-VL\,32B} \\
$i$ & CER & WER & CER & WER & CER & WER & CER & WER & CER & WER & CER & WER & CER & WER \\
\hline
 1 & 22.1 & 61.7 & 20.5 & 48.9 & 5.0 & 24.4 & \textbf{4.0} & \textbf{23.4} & 6.3 & 40.4 & 37.3 & 57.4 & 33.0 & 42.6 \\
 2 & 34.6 & 70.2 & 29.4 & 57.4 & 6.1 & 20.1 & \textbf{2.6} & \textbf{19.1} & 6.9 & 42.6 & 36.0 & 66.0 & 25.7 & 42.6 \\
 3 & 31.0 & 57.4 & 31.4 & 53.2 & 7.9 & \textbf{22.1} & 7.3 & 31.9 & \textbf{6.9} & 44.7 & 48.8 & 51.1 & 31.0 & 44.7 \\
 4 & 39.9 & 61.7 & 38.9 & 53.2 & 7.3 & \textbf{25.2} & \textbf{4.0} & 25.5 & 6.9 & 40.4 & 39.3 & 46.8 & 27.7 & 51.1 \\
 5 & 48.8 & 70.2 & 45.5 & 70.2 & 8.0 & \textbf{25.2} & 6.9 & 25.5 & \textbf{5.9} & 34.0 & 32.3 & 46.8 & 22.4 & 36.2 \\
 6 & 49.5 & 74.5 & 50.5 & 74.5 & 12.3 & 35.4 & 9.9 & 31.9 & \textbf{5.6} & \textbf{29.8} & 37.0 & 53.2 & 28.7 & 38.3 \\
 7 & 72.9 & 97.9 & 70.6 & 95.7 & 14.2 & 35.9 & 11.6 & \textbf{27.7} & \textbf{5.6} & 29.8 & 48.5 & 57.4 & 37.0 & 40.4 \\
 8 & 64.7 & 74.5 & 55.5 & 70.2 & 16.8 & 36.9 & 13.9 & \textbf{27.7} & \textbf{7.6} & 44.7 & 47.5 & 48.9 & 23.4 & 48.9 \\
 9 & 68.3 & 87.2 & 70.0 & 80.9 & 16.8 & 36.9 & 13.5 & \textbf{31.9} & \textbf{6.9} & 44.7 & 36.3 & 46.8 & 29.0 & 40.4 \\
10 & 81.2 & 117.0 & 79.5 & 106.4 & 16.9 & 37.4 & 14.2 & \textbf{36.2} & \textbf{5.9} & \textbf{36.2} & 37.0 & 51.1 & 24.8 & 42.6 \\
\hline
\end{tabular}}
\end{table}

\paragraph{CER vs.~WER for Gemini.} Gemini~3~Flash exhibits an
excellent CER ($\approx\!6\,\%$) but a markedly higher WER
($30$--$45\,\%$), systematically above Calfa~OCR (WER $19$--$36\,\%$).
This is not a recognition failure but a \emph{tokenization} effect:
the model normalises typography in ways that displace word boundaries
without altering many characters. Three recurring patterns account
for most of the gap: (i) spaces inserted around hyphens, e.g.
\textit{«\arm{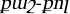}»} rendered \textit{«\arm{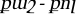}»} (one
token becomes three); (ii) case and diacritic normalisation in
out-of-language strings, e.g. \textit{«JAPHET FRÈRES»} $\rightarrow$
\textit{«JAPHET Frères»}; and (iii) the Armenian full-stop
[\textit{\arm{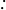}}] replaced by the colon [\textit{:}], breaking
line-final tokens. Each costs one or two characters but a whole word,
inflating WER while the transcription stays essentially correct. For
downstream reading-order and NLP, Gemini's CER is therefore the more
meaningful signal, and a light post-normalisation is applied before
evaluating WER. The gap is a formatting artefact, not a hallucination:
the check below finds no invented content in Gemini's output, so
post-normalisation is the applicable mitigation.

\newcommand{\thumbw}{0.085\linewidth}

\begin{figure*}[t]
  \centering
  \begin{tikzpicture}
    \begin{axis}[
        width=\linewidth, height=6cm,
        xlabel={Curvature intensity $i$ \;\;($\alpha,\beta,\gamma$ linearly scaled)},
        ylabel={CER},
        ymin=0, ymax=0.9,
        xmin=1, xmax=10,
        xtick={1,2,3,4,5,6,7,8,9,10},
        ytick={0,0.1,0.2,0.3,0.4,0.5,0.6,0.7,0.8},
        yticklabel={\pgfmathprintnumber[fixed,precision=2]{\tick}},
        grid=both, grid style={gray!20},
        legend columns=4,
        legend style={at={(0.5,1.03)}, anchor=south, font=\scriptsize,
                      fill=white, draw=gray!40,
                      /tikz/every even column/.append style={column sep=5pt}},
        legend cell align=left,
        tick label style={font=\small},
        label style={font=\small},
    ]
      \addplot[mark=square*, thick, red!70!black] coordinates {
        (1,0.221)(2,0.346)(3,0.310)(4,0.399)(5,0.488)
        (6,0.495)(7,0.729)(8,0.647)(9,0.683)(10,0.812)};
      \addlegendentry{Tesseract\,5 (default)}

      \addplot[mark=triangle*, thick, orange!90!black] coordinates {
        (1,0.205)(2,0.294)(3,0.314)(4,0.389)(5,0.455)
        (6,0.505)(7,0.706)(8,0.555)(9,0.700)(10,0.795)};
      \addlegendentry{Tesseract\,5 (Calfa)}

      \addplot[mark=*, thick, blue!70!black] coordinates {
        (1,0.040)(2,0.026)(3,0.073)(4,0.040)(5,0.069)
        (6,0.099)(7,0.116)(8,0.139)(9,0.135)(10,0.142)};
      \addlegendentry{Calfa\,OCR}

      \addplot[mark=star, thick, brown!80!black] coordinates {
        (1,0.050)(2,0.061)(3,0.079)(4,0.073)(5,0.080)
        (6,0.123)(7,0.142)(8,0.168)(9,0.168)(10,0.169)};
      \addlegendentry{CRAFT+Tess.\,(Calfa)}

      \addplot[mark=diamond*, thick, green!50!black] coordinates {
        (1,0.063)(2,0.069)(3,0.069)(4,0.069)(5,0.059)
        (6,0.056)(7,0.056)(8,0.076)(9,0.069)(10,0.059)};
      \addlegendentry{Gemini\,3 Flash}

      \addplot[mark=pentagon*, thick, violet!70!black] coordinates {
        (1,0.373)(2,0.360)(3,0.488)(4,0.393)(5,0.323)
        (6,0.370)(7,0.485)(8,0.475)(9,0.363)(10,0.370)};
      \addlegendentry{Qwen2.5-VL\,72B}

      \addplot[mark=oplus*, thick, cyan!70!black] coordinates {
        (1,0.330)(2,0.257)(3,0.310)(4,0.277)(5,0.224)
        (6,0.287)(7,0.370)(8,0.234)(9,0.290)(10,0.248)};
      \addlegendentry{Qwen3-VL\,32B}

      \draw[dashed, red!50, thick] (axis cs:5.5,0) -- (axis cs:5.5,0.9);
      \node[red!60!black, anchor=south west, font=\scriptsize\itshape]
        at (axis cs:5.55,0.78) {segmentation collapse};
    \end{axis}

    \node[anchor=north, inner sep=2pt] at (current bounding box.south)
      {%
        \foreach \i in {1,...,10}{%
          \includegraphics[width=\thumbw]%
            {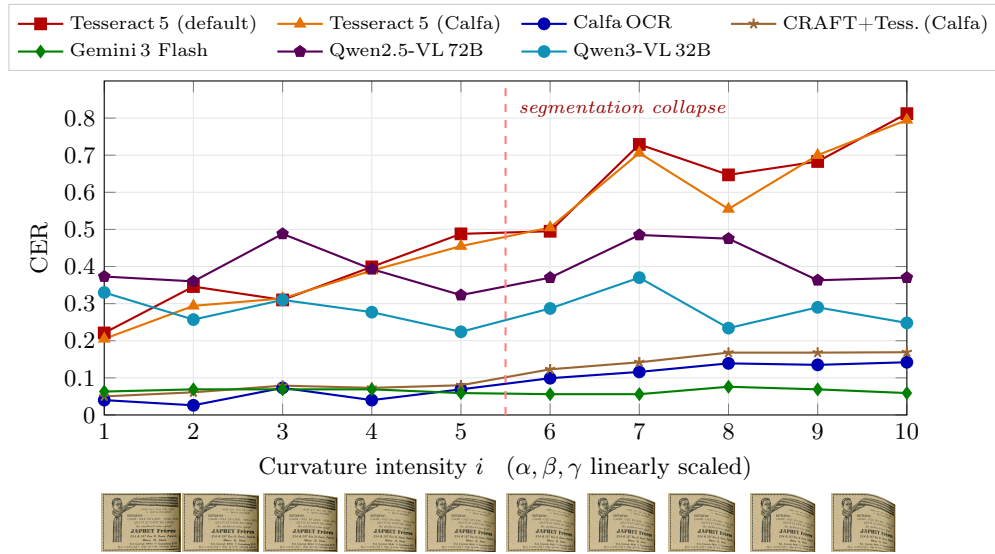}\hspace{1pt}%
        }%
      };
  \end{tikzpicture}
  \caption{Average CER of seven OCR systems on the advertisement test
    set. Intensity $i\!=\!1$ is the unaltered test set (general OCR
    baseline); $i\!=\!2,\dots,10$ apply the synthetic curvature
    described in Section~\ref{sec:ocr}. The dashed line marks the
    regime past which Tesseract's own page segmenter fails on the
    curved layout; replacing only its detector with CRAFT (brown)
    avoids the collapse with the same recognizer. The thumbnails
    below the axis show one representative advertisement of the test
    set at each intensity~$i$ (used only for illustration; the curves
    are computed over the whole test set).}
  \label{fig:curvature}
\end{figure*}

\paragraph{Qualitative behaviour of the open-weight VLMs.} The flat
error floor of Qwen2.5-VL and Qwen3-VL hides a script-dependent
split. Both models transcribe French/Latin-script lines (addresses,
shop names, phone numbers) almost perfectly, e.g.
\textit{«CACHE-COLS EN LAINE, PULL-OVER, GILETS ET GANTS DE LAINE»}
and \textit{«224 \& 257 Rue St. Denis. Paris 2e.»} are reproduced
verbatim or with only cosmetic spacing changes. On Armenian lines,
however, both models frequently drift into fluent but incorrect
Armenian rather than transcribing the actual glyphs: the
ground-truth \textit{«\arm{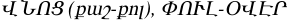}»} is rendered
as \textit{«\arm{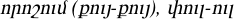}»} by Qwen2.5-VL,
while Qwen3-VL drops the line entirely for an unrelated word
(\textit{«\arm{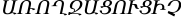}»}). Unlike Gemini's near-miss
tokenization errors, this output is semantically unrelated to the
source, consistent with weaker Armenian-script coverage in these
models' pretraining data: Qwen's CER is therefore not a reliable
proxy for usable transcription quality on this corpus, and neither
model is currently a viable substitute for Gemini~3~Flash or
Calfa~OCR on Armenian text.

\paragraph{Hallucination control for Gemini.} A generative OCR also
risks \emph{hallucinations}: invented addresses, fabricated telephone
numbers, or proper nouns drifting toward likely-looking alternatives.
Because CER alone cannot distinguish ``wrong character'' from
``correct character of an invented word'', Gemini is cross-checked
against the manually annotated advertisement test set with an
ad-level similarity score. For each advertisement detected by the layout model, Gemini
is queried for a structured record (Armenian/French title and
description, address and its components, telephone, languages,
topics; see Figure~\ref{fig:gemini_prompt} in the Appendix) and
each predicted description string $s_2$ is compared with its
best-matching manual description $s_1$ using a normalised
Levenshtein ratio
\[
  \rho(s_1,s_2) \;=\; 1 \;-\;
        \frac{\mathrm{lev}(s_1,s_2)}{|s_1|+|s_2|} \;\in\;[0,1],
\]
matched by mutual nearest neighbour (one manual description
$\leftrightarrow$ one Gemini description per ad). The bands are
read as $\rho\!\geq\!0.85$ near-exact, $0.70$--$0.84$ good match
with minor edits, $0.50$--$0.69$ partial, and $\rho\!<\!0.50$ poor.

Across the test set, $\rho$ concentrates between $0.78$ and $0.95$.
Manual inspection of every pair below $0.85$ shows that the
residual gap is \emph{not} fabricated content: Gemini routinely
transcribes information present in the image but \emph{absent from
the manual ground truth} -- in-frame parentheticals such as
\textit{«\arm{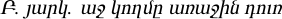}»} ("2nd floor, right
side, first door") account for most of the edit distance. No
hallucination was observed on the test set.
The check is therefore best read symmetrically -- low ratios flag
divergences for human review, and on this corpus those divergences
favour the model, arguing for augmenting the ground truth with
Gemini's reads before measuring information-extraction precision
downstream.

\section{Geolocation results, database and map}
\label{sec:geo}

Each extracted address is sent through the geocoding cascade of
Section~\ref{sec:pipeline} (Nominatim, then historical gazetteers
above similarity $0.5$). On a manually inspected sample of $95$
advertisements from the $1926$ run, $86$ ($\approx\!90.5\,\%$) show
no identifiable geocoding error against the source ad -- a small
manual sample used for failure-mode analysis, \emph{not} a
systematic ground-truth precision/recall benchmark. A quantitative
evaluation against a dedicated ground-truth address list does not
yet exist, hence the qualitative analysis below; building such a
reference subset is the most pressing follow-up of
Section~\ref{sec:conclusion}.

\paragraph{Qualitative failure modes.} On the same sample, $9$
geocoded addresses ($\approx\!9.5\,\%$) fall into one of the
recurring patterns below.
\begin{itemize}\itemsep1pt
\item \textbf{Armenian-script tokens in the address line}: street
  names (\arm{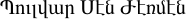} for \emph{Bd Saint-Germain}),
  city names (\arm{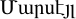} for \emph{Marseille}, where the
  cascade defaults to a Paris match and silently loses the
  Marseille branch), common words (\arm{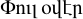} for
  \emph{pull-over}), branch indicators (\arm{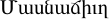}), and
  Armenian declensions affixed to French toponyms
  (\emph{Gare de l'Est}\arm{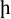}, genitive) all defeat the lookup.
\item \textbf{Multiple addresses in one ad} (Paris main shop
  plus suburban branch, several arrondissements, or different
  cities), geocoded on the first hit only; mixed-city cases are
  the worst -- ``\emph{120 Bd de Grenelle} (Paris); \emph{22 rue
  Nationale} (Billancourt)'' yields a single aberrant point.
\item \textbf{Typos in the original ad}, e.g.\ \emph{rue
  Lafayette} for \emph{rue La Fayette}.
\item \textbf{Gazetteer mismatches}: streets renamed since
  publication (``\emph{8--10 rue de l'Orient}'' is today
  \emph{rue de l'Armée d'Orient}); unexplained Nominatim drift
  (``\emph{2bis rue Conservatoire}'' resolves to Grenoble); Paris
  Time Machine's ranked similarity sometimes returns an unrelated
  street that merely shares a substring when the queried address
  no longer exists.
\item \textbf{Near-miss on the geocoded point}: correct street,
  wrong house number (``\emph{20 rue de Bezons, Courbevoie}''
  $\rightarrow$ no.~$7$), or correct neighbourhood with a
  neighbouring street in an adjacent commune
  (``\emph{30 Av.\ de Robinson, Sceaux-Robinson}'' $\rightarrow$
  \emph{Av.\ des 4 Chemins, Châtenay-Malabry}; rendered in
  Appendix~\ref{app:geocoding}).
\end{itemize}

The transliteration patterns of item~1 are largely systematic --
the same Armenian-script renderings recur from one ad to the next
for a small closed set of common French street, city and lexical
tokens -- and a dedicated Armenian-to-French equivalence
dictionary is being built to absorb them upstream of the cascade;
on the manually inspected sub-sample, this is expected to recover
most of the Armenian-script failures at near-zero runtime cost.
The same pre-processing step also splits each semicolon-separated
address against its own city, so that multi-city ads (item~2) are
geocoded marker by marker rather than collapsed onto a single
point.

Several near-misses (renamed streets, wrong number on the correct
street, neighbour-street in an adjacent commune) preserve the
\emph{locality} even when the precise point is wrong: for
anthropological studies of the Parisian Armenian diaspora the
neighbourhood signal already carries most of the information, so
``address-level'' vs.\ ``locality-level'' geocoding is an open
research direction to be settled with domain specialists. The
historical gazetteer used by the cascade
(SoDUCo~\cite{abadie2022benchmark,abadie2023soduco}) covers Paris
up to the 1920s and is being extended forward; once its coverage
reaches the 20th-century decades that dominate the corpus, this is
the most promising route to resolve the remaining ambiguous and
obsolete addresses.

\paragraph{Database.} The pipeline writes to a small SQLite store
with three tables: \texttt{images} (one row per \emph{Haratch} page:
identity, date, IIIF URL, status); \texttt{crops} (one row per
detected advertisement: YOLO box, Mistral / Gemini outputs, address
fields, IIIF crop URL); and \texttt{crop\_topics}, which normalises
the multi-valued taxonomy fields into $\langle$\texttt{type},
\texttt{value}$\rangle$ pairs. Each occurrence of a recurring ad is a
distinct \texttt{crops} row keyed by its source page, so a shop's
temporal frequency is preserved natively and aggregated downstream by
string- or address-level deduplication. The schema is intentionally
narrow: any consumer (map renderer, statistics, ML fine-tuning) reads
a single denormalised view of an ad with its tags.

\begin{figure}[t]
\centering
\frame{\includegraphics[width=\linewidth,keepaspectratio]{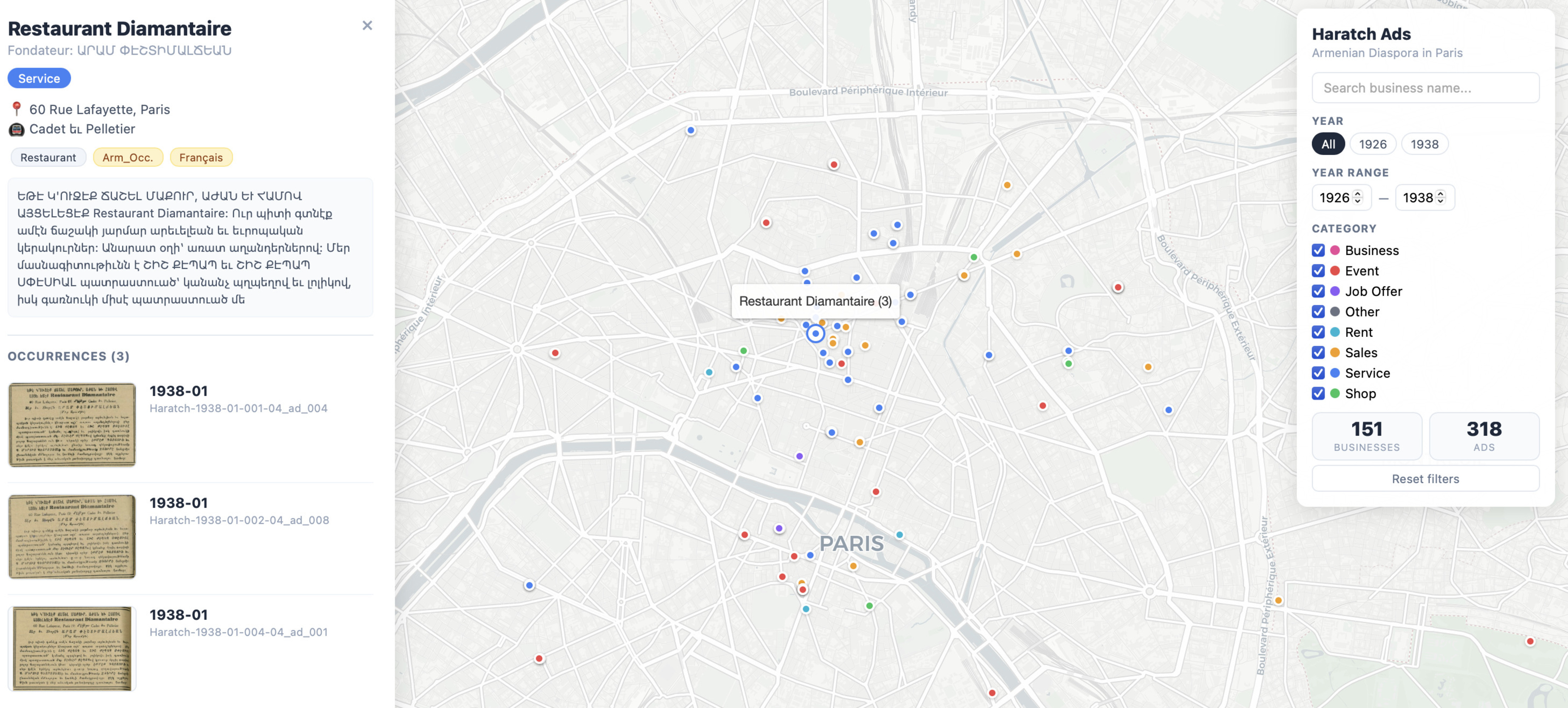}}
\caption{Interactive map of the geocoded advertisements (current
  prototype, modern OSM base). Year, language and category filters
  on the side panel; clicking a marker opens the IIIF dynamic crop
  of the originating page. POC on 1926, 1930 and 1938.}
\label{fig:map}
\end{figure}

\paragraph{Interactive map.} The geocoded crops are exposed on an
interactive web map (Figure~\ref{fig:map}) that overlays every
located advertisement on a modern Paris base, with filters on year,
language and category and a click-through to the IIIF crop of the
source page (the feedback loop of Figure~\ref{fig:pipeline}). For the
present proof of concept, $2{,}880$ pages covering the full $1926$,
$1930$ and $1938$ runs of \emph{Haratch} were processed, yielding
$961$ valid advertisements, of which $304$ are shown on the demo map
($148$ of them duplicates of the same business across issues,
attributed to a single entity in the database).

\section{Conclusion and perspectives}
\label{sec:conclusion}

MAP is an effective end-to-end pipeline: despite a minority of
missed borderless ads, it turns the Armenian press of France into a
navigable map of the 20th-century Parisian Armenian commercial
diaspora and opens a record until now inaccessible at scale to
anthropological and socio-historical study. On framed advertisements
the task is essentially solved -- the detector saturates and the
verifier wrongly drops only $1$ of $865$ crops ($0.1\,\%$) -- and the
residual work is concentrated on borderless blocks, which look like
ordinary editorial text and carry almost all of the recall cost.

Two findings stand out on the technical side. First, a VLM-based OCR
(Gemini) is essentially insensitive to the page-curvature regime that
collapses standard CRNN+CTC stacks, and is retained for the strongly
curved subset. A fully open CRAFT+Tesseract (Calfa) pipeline recovers most of
the same loss -- curvature is a detection, not a recognition,
bottleneck -- but the VLM remains the more robust of the two. This
robustness is a property of the model, not of VLMs in general:
open-weight alternatives read Latin-script lines well yet drift into
fluent, incorrect Armenian. Second, and more modestly, no single
accuracy metric separates the four closely-tied ad detectors, so
YOLOv12m is retained on efficiency grounds, with a few-shot
Mistral~Medium verifier on the borderless branch. Together, these
results show that VLM-driven data bootstrapping is a practical answer
to the under-resourced Armenian setting: it produces usable
structured records at scale on a corpus that no off-the-shelf layout
or OCR model covers today, and a companion HuggingFace demo lets
readers reproduce the full loop on their own crops.

\paragraph{Perspectives.} Short-term follow-ups include quantifying
the geocoding success rate against a manually verified subset,
replacing the modern OSM base with a 1920--1940 historical
cartography of Paris, opening the map to reader corrections for
addresses the cascade cannot resolve, and extending the
metadata-recovery branch to non-IIIF and metadata-incomplete sources
via a dedicated YOLO\,+\,Calfa~OCR/Gemini issue\,/\,year detector.
The principal next step is the production transition: fine-tuning a
locally-hosted multimodal VLM (e.g.\ Qwen with LoRA) to retire the
per-page API calls and scale to the several hundred thousand pages of
the full Armenian-press corpus.

\begin{credits}
\subsubsection{\ackname}
The Samuelian project was originally founded by the Calouste Gulbenkian Foundation. MAP is a BnF DataLab 2026 project, carried out in connection with the ANR DALiH project (ANR-21-CE38-0006), the HumaNum consortium Distam+, and the ANR DECIDON project (EPITA, ANR-25-CE38-4063). We would like to extend our particular thanks to the Samuelian family for their trust and for granting access to and use of the collection, as well as to Nicolas Samuelian for his thematic co-direction of this work.

\subsubsection{Code and dataset}
Code and models are available at: \url{https://huggingface.co/spaces/calfa-ai/mapping-armenian-paris}.
The Label Studio template and the $500$-page annotated corpus are
released at \url{https://calfa-co/map-project-public}.
\end{credits}

\bibliography{bibliography}

\begin{thebibliography}{10}
\providecommand{\url}[1]{\texttt{#1}}
\providecommand{\urlprefix}{URL }
\providecommand{\doi}[1]{https://doi.org/#1}

\bibitem{abadie2023soduco}
Abadie, N., Baciocchi, S., Bernard, C., Carlinet, E., Chapron, P., Chazalon,
  J., Chen, Y., Cristofoli, P., Dum{\'e}nieu, B., Fernandez, M., et~al.:
  Soduco: croisement de sources g{\'e}o-historiques pour l'{\'e}tude de
  l'{\'e}volution de paris de 1789 {\`a} 1950. In: Conf{\'e}rence
  interdisciplinaire: Extraction, traitement et visualisation de donn{\'e}es
  complexes en g{\'e}ographie (XVIIIe si{\`e}cle-XIXe si{\`e}cle) (2023)

\bibitem{abadie2022benchmark}
Abadie, N., Carlinet, E., Chazalon, J., Dum{\'e}nieu, B.: A benchmark of named
  entity recognition approaches in historical documents application to 19 th
  century french directories. In: International Workshop on Document Analysis
  Systems. pp. 445--460. Springer (2022)

\bibitem{baek2019character}
Baek, Y., Lee, B., Han, D., Yun, S., Lee, H.: Character region awareness for
  text detection. In: Proceedings of the IEEE/CVF conference on computer vision
  and pattern recognition. pp. 9365--9374 (2019)

\bibitem{duan2026glm}
Duan, S., Xue, Y., Wang, W., Su, Z., Liu, H., Yang, S., Gan, G., Wang, G.,
  Wang, Z., Yan, S., et~al.: Glm-ocr technical report. arXiv preprint
  arXiv:2603.10910  (2026)

\bibitem{fleischhacker2025enhancing}
Fleischhacker, D., Kern, R., G{\"o}derle, W.: Enhancing ocr in historical
  documents with complex layouts through machine learning. International
  Journal on Digital Libraries  \textbf{26}(1), ~3 (2025)

\bibitem{google2026gemini3flash}
{Google DeepMind}: Gemini 3 flash (2026),
  \url{https://aistudio.google.com/models/gemini-3}

\bibitem{gravier2024typology}
Gravier, J., Barthelemy, M.: A typology of activities over a century of urban
  growth. Nature Cities  \textbf{1}(9),  567--575 (2024)

\bibitem{grigoryan2025automated}
Grigoryan, A., Yeghiazaryan, A., Ispiryan, D., et~al.: {Automated Quality
  Assessment and Restoration of Scanned Armenian Newspapers}. Bulletin of
  Armenian Libraries  \textbf{8}(2),  34--43 (2025)

\bibitem{mermet2026galligeo}
Mermet, {\'E}., Grosso, E.: Galligeo: donner de l'espace aux cartes et plans de
  gallica. In: Journ{\'e}e annuelle des projets de recherche en r{\'e}sidence
  au BnF DataLab (2026)

\bibitem{mermet2024developing}
Mermet, {\'E}., Peignot, S., Rouet, P., Cador, H., Rinner, V., Varet-Vitu, A.:
  Developing a diachronic geohistorical repository: Paris 1900-2020--blocks,
  streets and addresses. Histoire urbaine  \textbf{70}(2),  39--64 (2024)

\bibitem{pinol2024adresses}
Pinol, J.L., Mermet, {\'E}.: Adresses, g{\'e}ocodage et information historique.
  Cartes \& G{\'e}omatique (255),  85--98 (2024)

\bibitem{semnani2025churro}
Semnani, S., Zhang, H., He, X., Tekg{\"u}rler, M., Lam, M.: Churro: Making
  history readable with an open-weight large vision-language model for
  high-accuracy, low-cost historical text recognition. In: Proceedings of the
  2025 Conference on Empirical Methods in Natural Language Processing. pp.
  34765--34812 (2025)

\bibitem{VidalGorene2023}
Vidal-Gorene, C.: {OCR / HTR technologies and Armenian Heritage Preservation}.
  Bulletin of Armenian Libraries pp. 61--65 (2023).
  \doi{10.52027/18294685-cvo2023.sp}

\bibitem{vidal2025armenian}
Vidal-Gor{\`e}ne, C., Decours-Perez, A., Kasparian, A., Tanelian, A., Ohanian,
  A.: Armenian htr: State of the art, transcription guidelines and good
  practices  (2025)

\bibitem{vidal2026under}
Vidal-Gor{\`e}ne, C., Kindt, B., Cafiero, F.: Under-resourced studies of
  under-resourced languages: lemmatization and pos-tagging with llm annotators
  for historical armenian, georgian, greek and syriac. In: Proceedings of the
  Second Workshop on Language Models for Low-Resource Languages (LoResLM 2026).
  pp. 324--334 (2026)

\bibitem{vidal2026-newspaper}
Vidal-Gor{\`e}ne, C., Tomeh, N., Khurshudyan, V.: {Semantic-Guided Reading
  Order Reconstruction in Historical Armenian Newspapers with LLMs}. In:
  {International Conference on Pattern Recognition}. Lyon, France (2026),
  \url{https://enc.hal.science/hal-05674556}

\bibitem{wei2025deepseek}
Wei, H., Sun, Y., Li, Y.: Deepseek-ocr: Contexts optical compression. arXiv
  preprint arXiv:2510.18234  (2025)

\bibitem{wei2026deepseek}
Wei, H., Sun, Y., Li, Y.: Deepseek-ocr 2: Visual causal flow. arXiv preprint
  arXiv:2601.20552  (2026)

\bibitem{xu2020layoutlm}
Xu, Y., Li, M., Cui, L., Huang, S., Wei, F., Zhou, M.: Layoutlm: Pre-training
  of text and layout for document image understanding. In: Proceedings of the
  26th ACM SIGKDD international conference on knowledge discovery \& data
  mining. pp. 1192--1200 (2020)

\bibitem{zhong2024opportunities}
Zhong, T., Yang, Z., Liu, Z., Zhang, R., You, W., Liu, Y., Sun, H., Pan, Y.,
  Li, Y., Zhou, Y., et~al.: Opportunities and challenges of large language
  models for low-resource languages in humanities research. arXiv preprint
  arXiv:2412.04497  (2024)

\end{thebibliography}

\appendix

\section{Appendice}
\label{app:prompts}

\begin{table}[htbp]
    \centering
    \caption{Per-region semantic schema. Conditional fields are revealed by the parent choice indicated in the description.}
    \label{tab:schema}
    \setlength{\tabcolsep}{4pt}
    \renewcommand{\arraystretch}{1.0}
    \resizebox{\linewidth}{!}{%
    \begin{tabular}{l l l p{4.8cm}}
    \hline
    Group & Field & Type & Values / description \\
    \hline
    Detection & \texttt{category} & rect.\ label & \texttt{with-frame}, \texttt{without-frame} \\
    \hline
    Offer
     & \texttt{off\_topic}   & text & Advertisement title \\
     & \texttt{off\_desc}    & text & Free-text description \\
     & \texttt{person\_name} & text & Proprietor / founder \\
    \hline
    Location
     & \texttt{address\_presence} & choice & \texttt{with\_address} / \texttt{without\_address} \\
     & \texttt{loc\_addr}, \texttt{loc\_dist}, \texttt{loc\_city}, \texttt{metro\_station} & text & Street, district / arrondissement, city, métro (when address present) \\
    \hline
    Contact & \texttt{telephone\_number} & text & Phone number as written \\
    \hline
    Languages & \texttt{ad\_lang} & choices & \texttt{Arm\_Occ.}, \texttt{Arm\_Or.}, \texttt{Français}, \texttt{English}, \texttt{Arm-Turc} \\
    \hline
    Categories
     & \texttt{ad\_topic} & choice & \texttt{Service}, \texttt{Job Offer}, \texttt{Sales}, \texttt{Event}, \texttt{Shop}, \texttt{Rent}, \texttt{Business}, \texttt{Other} \\
     & \texttt{ad\_topic\_service}  & choice & Medical, Restaurant, Hotel, Transport, Legal, Financial, Education, Tailor, Other (when \texttt{ad\_topic == Service}) \\
     & \texttt{ad\_topic\_business} & choice & Company, Startup, Agency, Other (when \texttt{== Business}) \\
     & \texttt{ad\_topic\_rent}     & choice & Apartment, House, Office, Other (when \texttt{== Rent}) \\
     & \texttt{ad\_topic\_job}      & choice & Education, Manual, Office, Other (when \texttt{== Job Offer}) \\
    \hline
    Visuals & \texttt{ad\_visuals} & choice & \texttt{Illustration}, \texttt{Logo}, \texttt{Frame} \\
    \hline
    \end{tabular}}
\end{table}

\begin{figure}[htbp]
    \centering
    \includegraphics[width=0.6\linewidth,height=0.23\textheight,keepaspectratio]{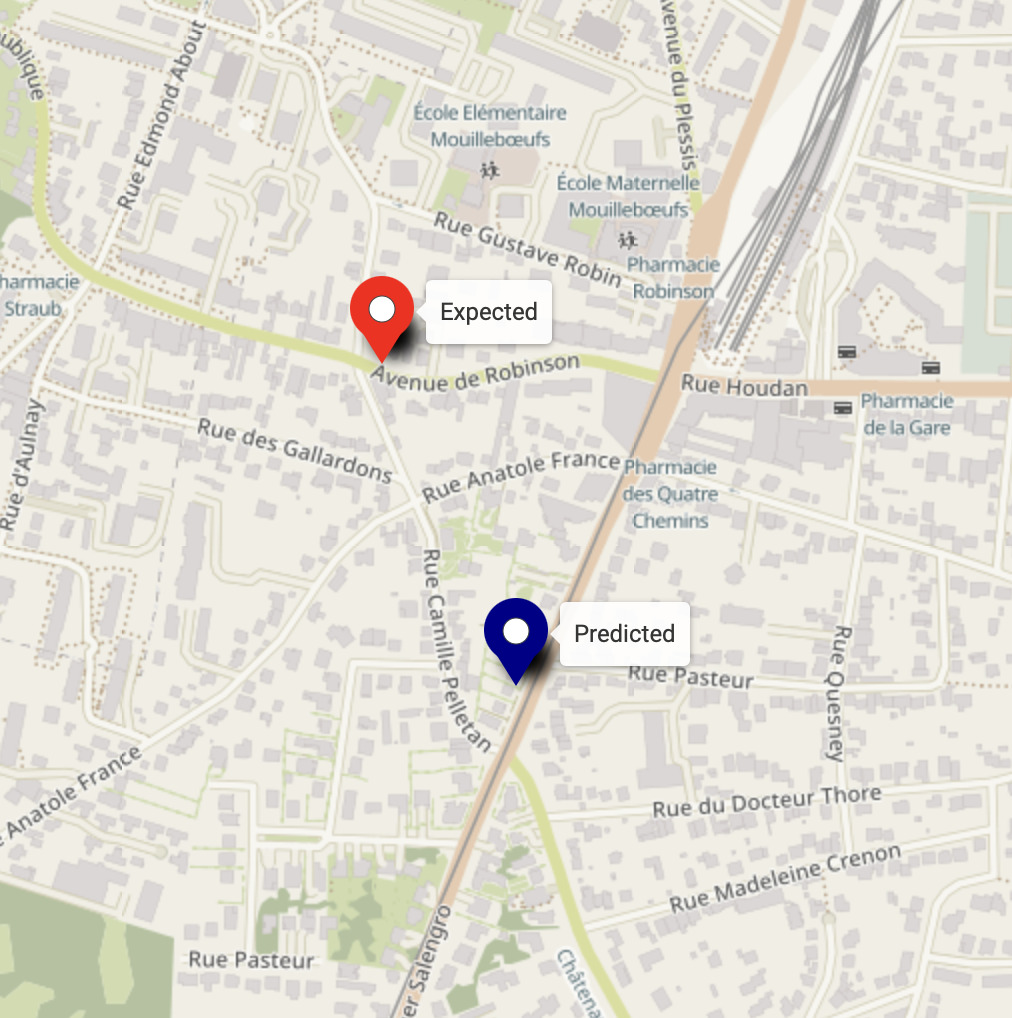}
    \caption{Map rendering of a near-miss geocoding case from the $1926$ sample (Section~\ref{sec:geo}): the ad's address ``\emph{30 Av.\ de Robinson, Sceaux-Robinson}'' is resolved to a neighbouring street in an adjacent commune (\emph{30 Av.\ des 4 Chemins, Châtenay-Malabry}, a few hundred metres away). The locality signal is preserved; the precise point is not (see Section~\ref{sec:geo} for discussion).}
    \label{app:geocoding}
\end{figure}

\begin{figure}[htbp]
    \centering
    \fbox{\begin{minipage}{0.93\linewidth}\scriptsize
    \textbf{Input.} A crop from a \emph{Haratch} (Paris) newspaper page.\par
    \textbf{Task.} Decide whether the crop is an \textsc{Advertisement}. Output a single word: \textsc{Yes} or \textsc{No}.\par
    \textbf{Early \textsc{No}.} If the crop contains \emph{Imprimerie}, \emph{Gérant}, \emph{Directeur}, \emph{Éditeur} or \emph{Rédacteur} (newspaper staff credits).\par
    \textbf{Strong \textsc{Yes} cues} (any one suffices). Street address (\emph{Rue}, \emph{Avenue}, \emph{Boulevard}, \emph{Place}); phone or \emph{Tél.}; Paris métro stop; arrondissement marker \texttt{(6e)}/\texttt{(9)}; opening hours; bold/large name block; decorative border; mixed Armenian/French; shop, service, restaurant, professional practice, hotel, pharmacy, bookstore; event (camp, concert, lecture, exposition); sale or rent of property, goods, clothing, books; job offer or job search; travel agency, shipping, transport.\par
    \textbf{Strong \textsc{No}} (override). Long continuous prose (article, editorial); birth / death / marriage notice; personal or family announcement.\par
    \textbf{Tie-break.} When in doubt, answer \textsc{Yes}.\par
    \textbf{Few-shot variant.} The same instructions above, prefixed
    with $7$ labelled example crops sent as image turns before the
    query crop: Ex.\,1--3 \textsc{Yes}/framed (visible ruled
    border), Ex.\,4--6 \textsc{Yes}/frameless (running text, no
    border), Ex.\,7 \textsc{No} (a non-advertisement crop: editorial
    text or a newspaper-internal notice) -- each followed by its
    expected verdict. The query crop is appended after the seven
    examples and answered exactly as in the zero-shot variant above.
    \end{minipage}}
    \caption{Mistral prompt for ad checking.}
    \label{fig:vlm_prompt}
\end{figure}

\begin{figure}[htbp]
  \footnotesize
    \centering
    \fbox{\begin{minipage}{0.93\linewidth}\scriptsize
    \textbf{Input.} An advertisement crop.\par\textbf{Task.} Extract structured fields without translating; return a JSON object.\par
    \textbf{Rules.} Extract text exactly as it appears in each language. \texttt{off\_title} is the business name -- not a person name, a product or a slogan. Do not duplicate Latin titles in both \texttt{\_arm} and \texttt{\_fr} unless both scripts appear; leave the other-language field empty when text appears in one language only. Multiple addresses are separated by semicolons; the $i$-th address corresponds to the $i$-th city.\par
    \textbf{Fields.} \texttt{off\_title\_arm}, \texttt{off\_title\_fr}, \texttt{founder}, \texttt{off\_desc\_arm}, \texttt{off\_desc\_fr}, \texttt{has\_frame}, \texttt{address\_presence}, \texttt{loc\_addr}, \texttt{loc\_dist}, \texttt{loc\_city}, \texttt{loc\_metro}, \texttt{telephone\_presence}, and the language / topic / sub-category enumerations of Table~\ref{tab:schema} (\texttt{ad\_lang}, \texttt{ad\_topic}, \texttt{ad\_topic\_service}, \texttt{ad\_shop\_type}, \texttt{ad\_topic\_event}, \texttt{ad\_topic\_rent}, \texttt{ad\_topic\_sales}, \texttt{ad\_topic\_business}, \texttt{ad\_topic\_job}). Numbers near \emph{Tel}, \emph{Téléphone}, \emph{Trudaine}, \emph{Nord Sud}, etc.\ count as phone numbers.\par
    \textbf{Output.} A JSON object with the keys above. Missing fields are empty strings, or empty lists for multi-valued fields.
    \end{minipage}}
    \caption{Gemini~3~Flash prompt used for the ad-level structured extraction.}
    \label{fig:gemini_prompt}
\end{figure}

\end{document}